%% file: main.tex
\documentclass[runningheads]{llncs}

\usepackage{eccv}

\usepackage{eccvabbrv}

\usepackage{hyperref}

\usepackage{orcidlink}

\input{setup/package}
\input{setup/macros}
\input{setup/symbols}

\input{setup/graphicspath}

\definecolor{urlcolor}{HTML}{D8127E}
\begin{document}

\title{Fast View Synthesis of Casual Videos with Soup-of-Planes} 

\titlerunning{Casual-FVS}

\author{
Yao-Chih Lee\inst{1,2*} \and
Zhoutong Zhang\inst{3} \and
Kevin Blackburn-Matzen\inst{2} \and \\
Simon Niklaus\inst{2} \and
Jianming Zhang\inst{2} \and
Jia-Bin Huang\inst{1} \and
Feng Liu\inst{2}
}

\authorrunning{Y.-C. Lee et al.}

\institute{$^1\ $University of Maryland College Park\hspace{5mm}$^2\ $Adobe Research\hspace{5mm}$^3\ $Adobe
}

\maketitle
\input{figures/fig1_teaser}


\input{0_abstract}
\input{1_introduction}
\input{2_related}

\input{3_method}

\input{4_result}
\input{5_conclusions}

%
%
\bibliographystyle{splncs04}
\bibliography{main}
\end{document}

%% file: setup/package.tex
\usepackage{graphicx}
\usepackage{subcaption}
\usepackage{float}
\usepackage{lscape}                                         
\usepackage{wrapfig}

\usepackage[lined,ruled,linesnumbered]{algorithm2e}
\usepackage{animate} 

\usepackage{booktabs}                   
\usepackage{multirow}

\usepackage{makecell}

\usepackage{paralist}
\usepackage{enumitem}
\setlist[enumerate]{itemsep=0mm}

\usepackage{bm}                          
\usepackage{epsfig}                      
\usepackage{mathtools}
\usepackage{amssymb,amsmath}   

\usepackage{units}
\usepackage{color}

\usepackage{comment}

\usepackage[hyphens]{url}
\usepackage{xspace}
\usepackage{setspace}
\usepackage{grfext}
\PrependGraphicsExtensions*{.jpg,.png,.PNG}

\usepackage[accsupp]{axessibility}  

\newcommand{\rulesep}{\unskip\ \vrule\ }

\usepackage[symbol]{footmisc}

\usepackage{soul} 
\usepackage{multirow}

\usepackage{pifont}

%% file: setup/macros.tex
\usepackage{soul}
\usepackage{svg}

\usepackage[accsupp]{axessibility} 

\usepackage[final]{changes}

\def\etal{et~al.}			  
\def\eg{e.g.,~}               
\def\ie{i.e.,~}               

\DeclareMathAlphabet{\altmathcal}{OMS}{cmsy}{m}{n}
\DeclareMathAlphabet{\mathbfit}{OT1}{ptm}{bx}{it}

\newlength\paramargin
\newlength\figmargin

\newlength\secmargin
\newlength\figcapmargin
\newlength\tabcapmargin

\newcommand{\topic}[1]
{
\vspace{1mm}\noindent\textbf{#1}
}

\long\def\ignorethis#1{}

\newbox\jsavebox%

\makeatletter
\newcommand{\providelength}[1]{%
  \@ifundefined{\expandafter\@gobble\string#1}
   {
    \typeout{\string\providelength: making new length \string#1}%
    \newlength{#1}%
   }
   {
    \sdaau@checkforlength{#1}%
   }%
}

\newcommand{\sdaau@checkforlength}[1]{%
  \edef\sdaau@temp{\expandafter\sdaau@getfive\meaning#1TTTTT$}%
  \ifx\sdaau@temp\sdaau@skipstring
    \typeout{\string\providelength: \string#1 already a length}%
  \else
    \@latex@error
      {\string#1 illegal in \string\providelength}
      {\string#1 is defined, but not with \string\newlength}%
  \fi
}
\def\sdaau@getfive#1#2#3#4#5#6${#1#2#3#4#5}
\edef\sdaau@skipstring{\string\skip}
\makeatother

\usepackage[capitalize]{cleveref}
\crefname{section}{Sec.}{Secs.}
\Crefname{section}{Section}{Sections}
\Crefname{table}{Table}{Tables}
\crefname{table}{Tab.}{Tabs.}

\newcolumntype{C}[1]{>{\centering\arraybackslash}p{#1}}
\newcolumntype{L}[1]{>{\raggedright\arraybackslash}p{#1}}
\newcolumntype{R}[1]{>{\raggedleft\arraybackslash}m{#1}}

\newcommand{\xmark}{\ding{55}}%

%% file: setup/symbols.tex
\def\xi{\mathbf{x}_i}

%% file: setup/graphicspath.tex
\graphicspath{{figures}, {examples}}

%% file: figures/fig1_teaser.tex
\makeatother
\begin{center}
    \centering
    \captionsetup{type=figure}
    \includegraphics[trim={2.4cm 4.7cm 0.5cm 3.5cm},clip,width=\linewidth]{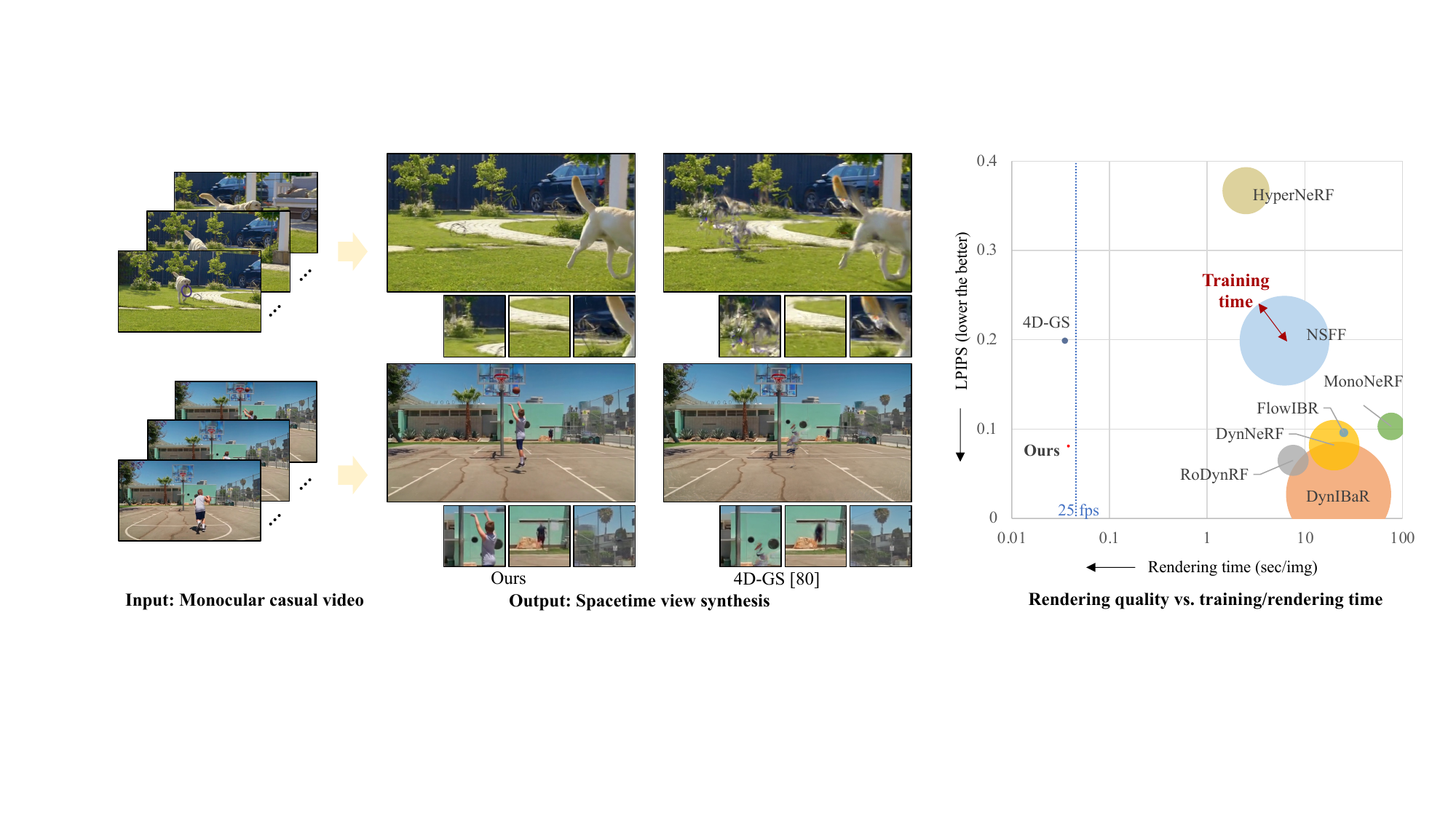}
    
    \setlength{\abovecaptionskip}{1ex}
    \setlength{\belowcaptionskip}{-0.5ex}
    \captionof{figure}{
    \textbf{Efficient dynamic novel view synthesis.}
    Our method only takes 15 minutes to optimize a representation from an in-the-wild video and can render novel views at 27 FPS. 
    On the NVIDIA Dataset~\cite{yoon2020nvidia}, our method achieves a rendering quality comparable to the state-of-the-art NeRF-based methods but is much faster to train and render. The bubble size in the figure indicates the training time (GPU-hours).
    }
    \label{fig:teaser}
\end{center}

%% file: 0_abstract.tex
\begin{abstract}

Novel view synthesis from an in-the-wild video is difficult due to challenges like scene dynamics and lack of parallax. While existing methods have shown promising results with implicit neural radiance fields, they are slow to train and render. This paper revisits explicit video representations to synthesize high-quality novel views from a monocular video efficiently. We treat static and dynamic video content separately. Specifically, we build a global static scene model using an extended plane-based scene representation to synthesize temporally coherent novel video. Our plane-based scene representation is augmented with spherical harmonics and displacement maps to capture view-dependent effects and model non-planar complex surface geometries. We opt to represent the dynamic content as per-frame point clouds for efficiency. While such representations are inconsistency-prone, minor temporal inconsistencies are perceptually masked due to motion. We develop a method to quickly estimate such a hybrid video representation and render novel views in real time. Our experiments show that our method can render high-quality novel views from an in-the-wild video with comparable quality to state-of-the-art methods while being 100$\times$ faster in training and enabling real-time rendering.
Project page at \textcolor{urlcolor}{\url{https://casual-fvs.github.io}}.
\keywords{Novel view synthesis \and casual video}

\footnotetext{*Work done while Yao-Chih was an intern at Adobe Research.}
\end{abstract}

%% file: 1_introduction.tex
\input{figures/fig2_3dgs_failure}
\section{Introduction}
\label{sec:intro}

Neural radiance fields (NeRFs)~\cite{mildenhall2020nerf} have brought great success to novel view synthesis of in-the-wild videos. Existing NeRF-based dynamic view synthesis approaches~\cite{nsff,gao2021dynerf,robustdynrf} rely on per-scene training to obtain high-quality results. However, the use of NeRFs as video representations makes the training process slow, often taking one or more days. 
Moreover, it remains challenging to achieve real-time rendering with such NeRF-based representations. 

Recently, 3D Gaussian Splatting~\cite{3dgaussian} based on an explicit scene representation achieves decent rendering quality on static scenes with a few minutes of per-scene training and real-time rendering. However, the success of 3D Gaussians relies on sufficient supervision signals from a wide range of multiple views, which is often lacking in monocular videos. As a result, floaters and artifacts are revealed in novel views in regions with a weak parallax (Fig.~\ref{fig:motivation_3dgs_failure}).

We also adopt a per-video optimization strategy to support high-quality view synthesis. Meanwhile, we seek a good representation for in-the-wild videos that is fast to train, allows for real-time rendering, and generates high-quality novel views. We use a hybrid representation that treats static and dynamic video content differently to handle scene dynamics and weak parallax simultaneously. We revisit plane-based scene representations, which are not only inherently friendly for scenes with low parallax but are effective at modeling static scenes in general. 
A good example is multi-plane image~\cite{zhou2018stereo,srinivasan2019pushing,flynn2019deepview,singlempi}. We, inspired by \textit{Piecewise Planar Stereo}~\cite{sinha2009piecewise}, use \textit{a soup of 3D oriented planes} to more flexibly represent the static video content from a wide range of viewpoints. To support temporally consistent novel view synthesis, we build a global plane-based representation for static video content. Moreover, we extend this soup-of-planes representation with spherical harmonics and displacement maps to capture view-dependent effects and complex non-planar surface geometry. Dynamic content in an in-the-wild video is often close to the camera and with complex motion. It is inefficient to maintain a large number of small planes to represent such content.
Consequently, we opt for per-frame point clouds to represent dynamic content for efficiency. To synthesize temporally coherent dynamic content and reduce occlusion, we blend the dynamic content from neighboring time steps. While such an approach is still inherently prone to temporal issues, small inconsistencies are usually not perceptually noticeable due to motion.

We further develop a method and a set of loss functions to optimize our hybrid video representation from a monocular video. Since our hybrid representation can be rendered in real-time, our per-video optimization only takes 15 minutes on a single GPU. Our method achieves a rendering quality that is comparable to NeRF-based dynamic synthesis algorithms~\cite{gao2021dynerf,nsff,robustdynrf,dynibar} quantitatively and qualitatively, but is over 100\texttimes\ faster for both training and rendering.




In summary, our contributions include:
\begin{itemize}
    \item a hybrid explicit non-neural representation that can model both static and dynamic video content, supports view-dependent effects and complex surface geometries, and enables real-time rendering;
    \item a per-video optimization algorithm with a set of carefully designed loss functions to estimate the hybrid video representation from a monocular video;
    \item extensive evaluations on the NVIDIA~\cite{yoon2020nvidia} and DAVIS datasets~\cite{davis} show that our method can generate novel views with comparable quality to SOTA NeRF-based methods while being 100\texttimes\ faster\footnotemark\  for training and rendering.
\end{itemize}

\footnotetext{\label{footnote_training_time}The training time does not include SfM preprocessing time (\eg COLMAP~\cite{colmap} or video-depth-pose estimation) for all methods.}


%% file: figures/fig2_3dgs_failure.tex
\begin{figure}
\centering
\includegraphics[trim={0.1cm 13cm 5.5cm 0.1cm},clip,width=\linewidth]
{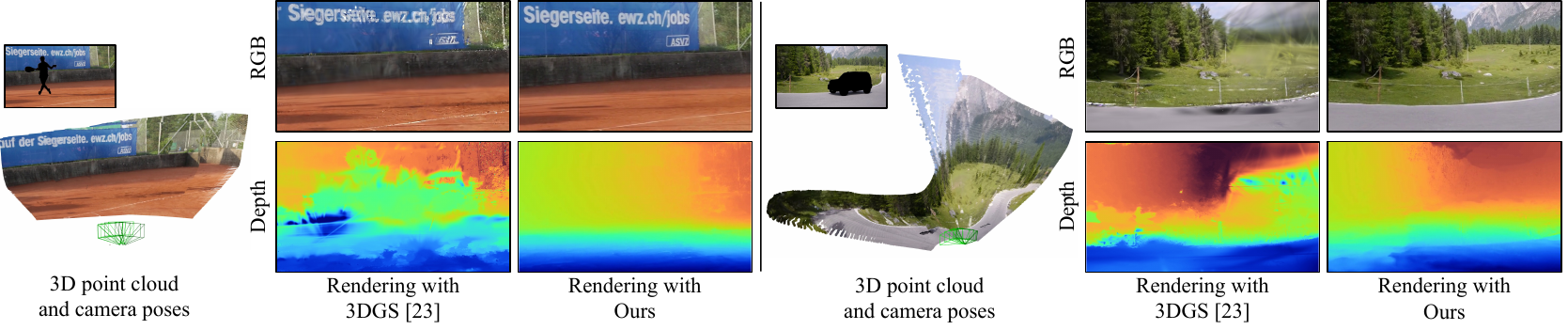}

\setlength{\abovecaptionskip}{0cm}
\setlength{\belowcaptionskip}{-0.6cm}
\caption{\textbf{3DGS~\cite{3dgaussian} fails in weak-parallax videos.} 
We show two casual videos in DAVIS~\cite{davis} that the 3D reconstructions show the weak parallax scenes by only camera rotations.
We use the ground truth masks to filter out the dynamics and only reconstruct the static scenes. We utilize the same 3D point cloud from video depth as the initialization for 3DGS and our method.
3DGS cannot handle such casual videos and produce floaters and noises in novel views due to insufficient parallax cues. 
}
\label{fig:motivation_3dgs_failure}
\end{figure}

%% file: 2_related.tex
\section{Related Work}
\label{sec:related}


\subsection{Dynamic-scene view synthesis}
In contrast to static-scene novel view synthesis~\cite{gortler1996lumigraph,levoy1996light,Shade1998,mildenhall2019llff,mildenhall2020nerf,martin2021nerfw,lin2021barf,wang2021ibrnet,suhail2022light,bian2023nope,meuleman2023localrf}, 
novel view synthesis for dynamic scenes is particularly challenging due to the temporally varying contents that need to be handled. To make this problem more tractable, many existing methods~\cite{zitnick2004high,stich2008view,Bemana2020xfields,bansal20204d,3dmaskvolume,dnerf2021,li2022neural3d,hyperreel,kplanes,hexplane,lin2023im4d,dynamic3dgaussian} reconstruct 4D scenes from multiple cameras capturing the dynamic scene simultaneously. However, such multi-view videos are not practical for casual applications, which instead only provide monocular videos.
To tackle monocular videos, Yoon~\etal~\cite{yoon2020nvidia} computes the video depth and performs depth-based 3D warping. However, the video depth may not be globally consistent, which results in view inconsistencies.

With the emergence of powerful neural rendering, a 4D scene can be implicitly represented in a neural network~\cite{xian2021space}. To model motion in neural representations,
~\cite{tretschk2021nrnerf,gao2021dynerf,nsff,nerfies,hypernerf,nerfplayer,robustdynrf} learn a canonical template with a deformation field to advect the casting rays. 
Some algorithms~\cite{gao2021dynerf,nsff,robustdynrf} utilize scene flow as a regularization for the 4D scene reconstruction which yields promising improvements.
Instead of embedding a 4D scene within the network parameters, DynIBaR~\cite{dynibar} aggregates the features from nearby views by a neural motion field to condition the neural rendering. 
Although existing neural rendering methods can achieve decent rendering quality, the computational costs and time for both per-scene training and rendering are high.
Recently, 
generalizable approaches~\cite{tian2023mononerf,busching2023flowibr,zhao2024pgdvs} 
investigated priors in the form of pre-training on a large corpus of data to reduce the per-scene training time for neural rendering. While showing encouraging results, these approaches do not generalize well though and the time-consuming rendering is still present.

\subsection{View synthesis with explicit representations}
In contrast to implicitly encoding a scene in the network parameters, view synthesis algorithms with explicit 3D representations can often train and/or render faster. 
Some methods exploit depth estimation to perform explicit 3D warping for novel view synthesis~\cite{3dkenburns,synsin,yoon2020nvidia,lee20213d,Cao2022FWD}. A feature-based point cloud is often used by neural rendering to enhance the synthesis quality~\cite{neuralpointgraphics,synsin,Cao2022FWD,Rockwell2021pixelsynth}. 
Instead of learning a feature space for point clouds, NPC~\cite{neuralpixelcompo} directly renders the RGB points with an MLP which yields a fast convergence and promising quality. 
Recent 3D/4D Gaussian approaches~\cite{3dgaussian,wu4dgaussians} treat each 3D point as an anisotropic 3D Gaussian to learn and render high-quality novel views efficiently without neural rendering. Nevertheless, these methods heavily rely on accurate point locations for a global 3D point cloud. Therefore, they may require depth sensors~\cite{neuralpointgraphics} or SfM~\cite{colmap} as initialization~\cite{neuralpixelcompo,3dgaussian}. 
Based on an initial point cloud, subsequent approaches~\cite{3dgaussian} leverage multi-view supervision to adaptively densify and prune the 3D points. However, these methods often fail in scenes with little parallax  (Fig.~\ref{fig:motivation_3dgs_failure}). Because of this, recent 4D Gaussians methods~\cite{dynamic3dgaussian,wu4dgaussians,yang2023deformable,yang2023gs4d,katsumata2023efficient,kratimenos2024dynmf,huang2023sc,das2023neural,lin2023gaussian,liang2023gaufre,li2023spacetime} are still limited to the setting with \emph{multi-view input videos} or \emph{quasi-static scenes} with large view angle changes for strong parallax cues.

Meshes are also a popular 3D representation~\cite{hu2021worldsheet}. However, it is difficult to estimate meshes from a casual video. One particular challenge is to directly optimize the positions of mesh vertices to integrate inconsistent depths from multiple views into a global mesh. 
Alternatively, \cite{wang2021neus,neus2,ren2023volrecon,yariv2023bakedsdf} first learn a neural SDF representation and then bake an explicit global mesh for static scenes. However, such a two-step method which involves optimizing an MLP is slow.

Layered depth images (LDI)~\cite{Shade1998} are an efficient representation for novel view synthesis~\cite{Shih3DP20,Kopf-OneShot-2020,li2023_3dcinemagraphy}. Multiplane image (MPI) approaches~\cite{zhou2018stereo,srinivasan2019pushing,singlempi,flynn2019deepview,wizadwongsa2021nex,3dmaskvolume,peng2022mpib,adampi}, further extend the LDI representation and use a set of fronto-parallel RGBA planes to represent a static scene. These MPIs can often be generated using a feed-forward network and are thus fast to estimate. They can also be rendered efficiently by homography-based warping and alpha composition. However, fronto-parallel planes are restricted to forward-facing scenes and do not allow for large viewpoint changes for novel view synthesis. 
To address this issue, 
\cite{lin2022neurmips,zhang2023structural} construct a set of oriented feature planes to perform neural rendering for static-scene view synthesis. Yet again however, such feature planes require a time-consuming optimization. Our method, similarly inspired by~\cite{sinha2009piecewise}, fits a soup of oriented planes to 3D scene surfaces. In contrast to feature planes, we adopt the non-neural RGBA representation in \cite{zhou2018stereo} for fast training and rendering.


%% file: 3_method.tex

\def\D{\mathcal{D}}  
\def\I{\mathcal{I}}  
\def\m{\mathcal{M}}            
\def\M{\mathcal{M}}  
\def\C{\mathcal{C}}  
\def\F{\mathcal{F}}  
\def\a{\alpha}       
\def\P{P}            
\def\pose{\pi}            
\def\displace{\Delta}
\def\l{l}  
\def\H{H} 
\newcommand{\nv}[1]{\tilde{#1}} 

\def\n{\mathbf{n}} 
\def\x{\mathbf{x}} 
\def\X{\mathbf{X}} 
\def\v{\mathbf{v}} 
\def\e{\epsilon}

\def\L{\mathcal{L}} 

\def\ng{\hspace{-0.1mm}}
\def\neg{\hspace{-0.2mm}}
\def\pos{\hspace{0.2mm}}


\input{figures/fig3_method_overview}
\section{Method}
\label{sec:method}

As illustrated in Fig.~\ref{fig:method_overview}, our method takes an $T$-frame RGB video, $\I_{1..T}$ as an input and renders a novel view $\nv{\I}_t$ at the target view point $\pose_{\mathrm{target}}$ and timestamp $t$. We first preprocess the input video to the obtain video depth maps $\D_{1..T}$, the camera trajectory $\pose_{1..T}$, and dynamic masks $\M_{1..T}$(Sec.~\ref{sec:preprocessing}). We then decompose the video into a global static representation (Sec.~\ref{sec:static_model}) and a per-frame dynamic representation (Sec.~\ref{sec:dynamic_model}). Finally, we render the static and dynamic representations according to the target camera pose and composite them to generate the novel view $\nv{\I}_t$.

We aim for a novel view synthesis approach that can train fast, support real-time rendering, and generate high-quality and temporally coherent novel views. As neural scene representations require more computation, we revisit explicit scene representations for a monocular video. First, we represent the dynamic content and the static background separately. We use a global static scene representation to enable temporally coherent view synthesis. To cope with dynamic content, we estimate a per-frame representation. 
While this is not ideal, minor inconsistencies within the dynamic content are not noticeable to viewers due to the motion-masking effect of human perception. Second, we use a soup of plane representation, inspired by \textit{Piecewise Planar Stereo}~\cite{sinha2009piecewise}, to represent the background and further extend it to support view-dependent effects and non-planar scene surfaces. Third, we represent dynamic content using per-frame point clouds. As detailed later in this section, we provide a method that can efficiently estimate such a hybrid video representation to support real-time rendering of with comparable quality of SOTA methods that need 100\texttimes\ of our training time.

\subsection{Preprocessing}
\label{sec:preprocessing}

Similar to existing methods~\cite{gao2021dynerf,nsff,dynibar}, our method obtains an initial 3D reconstruction from an input video $\I_{1..T}$ using off-the-shelf video depth and pose estimation methods. 
Specifically, we use a re-implementation of RCVD~\cite{robustcvd} 
by default and can also work on CasualSAM~\cite{casualsam}
to acquire video depth $\D_{1..T}$ and camera poses $\pose_{1..T}$. 
To obtain initial masks for dynamic regions, we first estimate dynamic regions through semantic segmentation~\cite{maskrcnn} and then acquire binary motion masks by thresholding the error between optical flows~\cite{raft} and rigid flows computed from depth maps and pose estimates. 
We then aggregate these masks to obtain the desired dynamic masks $\M_{1..T}$ before using Segment-Anything~\cite{segmentanything} to refine the object boundaries.
We find that this simple strategy works well for a wide variety of videos.

\subsection{Extended Soup of Planes for Static Content}
\label{sec:static_model}

We fit a soup of oriented planes to the point cloud constructed using the pre-computed depth maps and camera poses in Sec.~\ref{sec:preprocessing}. 
Each plane has the same texture resolution, which contains an appearance map and a density map to represent scene surfaces. 
We further augment the planes with spherical harmonics and displacement fields to model view-dependent effects and non-planar surfaces.

\topic{Plane initialization.} Given a number $N$ of planes $\{\P_i\}_{i=1}^N$, we first fit them to the 3D static scene point cloud by minimizing the objective:
\begin{equation}
    \Sigma_{i=1}^{N} d(\P_i, \X_{i,j}) + \lambda_{norm} \langle \n_{\P_i}, \n_{\X_{i,j}} \rangle + \lambda_{area} w_i h_i,
\end{equation}
where each 3D point $\X_j$ is assigned to the nearest plane $\P_i$ with the size $(w_i, h_i)$ in every optimizing iteration. 
The point-to-plane distance is calculated by {\scriptsize$
    d(\P, \X)=\sqrt{\max(|x^P| - \frac{w}{2}, 0)^2 + \max(|y^P|-\frac{h}{2}, 0)^2 + |z^P|^2}, 
$}
where {\scriptsize $[x^P, y^P, z^P]^\intercal$} is the 3D point $\X$ w.r.t. the plane basis coordinate system with the plane center as the origin. The optimizing variables $\theta^{\mathrm{plane}}$ include the planes' centers, widths, heights, and basis vectors. We also measure the orientation difference between the point normal vector $\n_{\X_{i,j}}$ and the plane normal $\n_{\P_i}$, and encourage compact plane size to avoid redundant overlapping. $\lambda_{norm}$, and $\lambda_{area}$ are weighting hyper-parameters. 
We present more details in the supplementary material.

\input{figures/fig4_view-dependent_texture}
\topic{View-dependent plane textures.} 
A 2D plane texture stores $S\times S$ appearance $\C_i$ and transparency maps $\a_i$ of the corresponding 3D plane $\P_i$. We utilize spherical harmonic (SH) coefficients for appearance maps to facilitate view dependency~\cite{Ramamoorthi2001,wizadwongsa2021nex,3dgaussian}. The view-specific color is computed as $\C^{\v} = \sum_{\ell=0}^{\ell^\C_{\max}}\C^\ell H^\ell(\v),$ where
$H^\ell$ represents the SH basis functions. 
Nonetheless, a flat plane with a view-dependent appearance map may still inadequately represent a bumpy surface (Fig.~\ref{fig:view-dependent-texture}a). 
The different viewing rays looking at the same 3D point may hit the plane in different locations, which can cause blurriness. 
With only view-dependent colors, the view-independent transparency $\a$ still cannot represent the geometry of the actual bumpy surface well. However, simply enabling a view-dependent transparency could introduce instability to the scene optimization.
To mitigate this issue, we introduce a \textit{view-dependent displacement} $\displace_i$ for each plane $\P_i$, encoded by SH coefficients.
This displacement map enables the adjustment of RGBA values from their original locations to specific locations based on the queried view directions.
As in Fig.~\ref{fig:view-dependent-texture}b, 
given a ray $\v$, we obtain the view-specific displacement, $\displace^{\v}= \sum_{\ell=0}^{\ell^\displace_{\max}}\displace^\ell H^\ell(\v)$, to shift the color $\C^{\v}$ into $\C^{\v,\displace^{\v}}$:
\begin{equation}
    \C^{\v,\displace^{\v}}(u, v) = \C^\v(u + \displace^{\v}_u, v + \displace^{\v}_v),
\end{equation}
where $(u, v)$ denotes the pixel on a plane that  ray $\v$ hits.
We also apply the displacement to the transparency map $\a$ along with color $\C^{\v}$, allowing a plane to better approximate complex non-planar surface geometries.

\topic{Differentiable rendering.}
We first obtain the view-specific RGB $\{\C^{\v,\Delta^\v}_i\}_{i=1}^N$ and transparency maps $\{\a^{\Delta^\v}_i\}_{i=1}^N$ from the texture of each plane $\P_i$. We then backward warp them to the target view $\pose_{\mathrm{target}}$ as $\{\nv{\C}^{\v,\Delta^\v}_i, \nv{\a}^{\Delta^{\v}}_i\}_{i=1}^N$ using planar homography $\{\H_i\}_{i=1}^N$ and composite them from back to front~\cite{zhou2018stereo}. 
Unlike fronto-parallel MPIs that have a set of planes with fixed depth order~\cite{zhou2018stereo}, we need to perform pixel-wise depth sorting to the unordered and oriented 3D planes before compositing them into the static-content image $\nv\I^s$:
\begin{equation}
\nv{\I}^s = \sum^{N}_{i=1} \left(\nv{\C}^{\v,\Delta^\v}_i\nv{\a}^{\Delta^{\v}}_i\prod^{N}_{j=i+1}(1-\nv{\a}^{\Delta^{\v}}_j)\right).
\end{equation}
Similarly, the static depth $\nv{\D}^s$ can be obtained by replacing color $\nv{\C}^{\v,\Delta^\v}_i$ with plane depth $\nv{d}_i$ in the above equation.

\subsection{Consistent Dynamic Content Synthesis}
\label{sec:dynamic_model}

Using oriented planes to represent near and complex dynamic objects is challenging. Therefore, we resort to simple but effective per-frame point clouds to represent them.
At each timestamp $t$, we extract the dynamic appearance $\I^{d}_t$ and the learned dynamic mask $\M^*_t$ from input frame $\I_t$ (Sec.~\ref{sec:optimization}). Then, they are warped to compute the dynamic color $\nv{\I}^d_{t}$ and mask $\nv{\M}_{t}$ at the target view $(\pose_{\mathrm{target}}, t)$ using forward splatting:
\begin{equation}
    \nv{\x}_t = K_{\mathrm{target}}\pose_{\mathrm{target}}\pose_t^{-1}\D_{t}K_{t}^{-1}\x_{t},
\end{equation}
where $\x_{t}$ and $\nv{\x}_t$ are pixel coordinates in the source $\I^d_{t}$ and target image $\nv{\I}_t$, respectively. $K$ are the camera intrinsics. We adopt differentiable and depth-ordered softmax-splatting~\cite{softmaxsplat,niklaus2023splatting} to warp $(\I^d_t, M^*_t)$ to $(\nv{\I}^d_t, \nv{M}_t)$ as well as the dynamic depth $\nv{D}^d_t$ w.r.t. view $\pose_{\mathrm{target}}$. The final image $\nv\I_t$ at $(\pose_{\mathrm{target}}, t)$ is a blend of the static $\nv{\I}^s$ and dynamic $\nv\I^d_t$:
\begin{equation}
\nv\I_t = (1-\nv\M'_t) \nv{\I}^s + \nv\M'_t \nv\I^d_t,
\end{equation}
where the soft mask $\nv\M'_t$ is based on the warped mask $\nv\M_t$ and further considers the depth order between $\nv\D^s_t$ and $\nv\D^d_t$ to handle occlusions between them. Similarly, the final depth $\nv\D_t$ can be computed from $\nv\D^s_t$ and $\nv\D^d_t$.

\topic{Temporal neighbor blending.}
Ideally, the dynamic appearance $\I^d_t$ and learned mask $\M^*_t$ can be optimized from the precomputed mask $\m_t$. But the precomputed mask $\m_t$ may be noisy and its boundary may be temporally inconsistent. As a result, extracting masks $\M^*_{1..T}$ independently from the noisy precomputed $\m_{1..T}$ can result in temporal inconsistencies.
Therefore, we sample and blend the dynamic colors and masks from neighboring views $\I^d_{t\pm j}$ with $\I^d_{t}$ via the optical flow $\F_{t\to t\pm j}$~\cite{raft}. The blended dynamic color $\overline{\I}^d_t$ and mask $\overline{\M}_t$ are then warped prior to compositing them on top of the static content. 


\subsection{Optimization}
\label{sec:optimization}
\topic{Variables.} 
We jointly optimize our hybrid static and dynamic video representation. For static content, in addition to the plane textures $\{\C_{i}, \a_{i}, \Delta_{i}\}_{i=1}^N$,  plane geometries $\{\theta_i^{\mathrm{plane}}\}_{i=1}^N$ (\ie plane basis, center, width, and height), and the precomputed camera poses $\pose_{1..T}$ can also be optimized. For dynamic content, we first initialize the RGB and masks $(\I^d_{1..T}, \M^*_{1..T})$ from the input frames $\I_{1..T}$ and precomputed masks $\M_{1..T}$ and then optimize them. Besides, we also refine flow $\F$ by fine-tuning flow model~\cite{raft} for neighbor blending during the optimization. 
We also optimize the depth $\D^d_{1..T}$ for dynamic content when scene flow regularization is adopted.
Then, we employ a recipe of reconstruction objectives and regularizations to assist optimization.

\topic{Photometric loss.}
The main supervision signal is the photometric difference between the rendered view $\nv\I_t$ and the input frame $\I_t$ at view $\pose_{\mathrm{target}}$. We omit time $t$ in this section for simplicity. The photometric loss $\L_{pho}$ is calculated as:
\begin{equation}
    \L_{pho}(\nv\I, \I) = (1-\gamma)\|\nv\I - \I\|_2^2 + \gamma \text{DSSIM}(\nv\I, \I),
\end{equation}
where DSSIM is the structural dissimilarity loss based on the SSIM metric~\cite{ssim} with $\gamma=0.2$. Besides, the perceptual difference $\L_{pho}^{percep}$~\cite{johnson2016perceptual} is measured by a pretrained VGG16 encoder.
Furthermore, to ensure the static planes represent static contents without the dynamic representation picking any static content, 
we directly compute the photometric loss in static regions between $\nv\I^s$ and $\I$ by:
\begin{equation}\label{eq:masked_photometric}
    \L_{pho}^s = \min\left(\L_{pho}^{\m}(\nv\I^s, \I),\L_{pho}^{\M^*}(\nv\I^s, \I)\right).
\end{equation}
\topic{Dynamic mask.}
The soft dynamic mask $\M^*$ blends the dynamic $\I^d$ with the static content. We compute a cross entropy loss $\L_{mask}^{bce}$ between $\M^*$ and the precomputed mask $\m$ with a decreasing weight since $\m$ may be noisy. 
We also encourage the smoothness of $\M^*$ by an edge-aware smoothness loss $\L_{mask}^{smooth}$~\cite{godard2017monodepth}.
To prevent the mask from picking static content, we apply a sparsity loss, $\L_{mask}^{reg}$, with both $L_0$- and $L_1$-regularizations~\cite{omnimatte} to restrict non-zero areas:
\begin{equation}
\L_{mask}^{reg}(\M^*)=
    \mu_{0}\Phi_0(\M^*) 
    + \mu_{1} \|\M^*\|_1  
    + \mu_{bce} L_{bce}(\M^*, \mathbf{1}),
\end{equation}
where $\Phi_0(\cdot)$ is an approximate $L_0$~\cite{omnimatte}. For non-zero areas, we encourage them to be close to 1 via the binary cross entropy $L_{bce}(\cdot)$ with a small weight.

\topic{Depth alignment.}
We use a depth loss to maintain the geometry prior in the precomputed $\D$ for the rendered depth $\nv\D$ by $\L_{depth} = \|\nv\D - \D\|_1 / |\nv\D + \D|$. Since the static depth $\nv\D^s$ should align with the depth used for warping dynamic content for consistent static-and-dynamic view synthesis, 
we measure the error between static depth $\nv\D^s$ and $\D$ similarly to the masked photometric loss in Eq.~\ref{eq:masked_photometric}:
\begin{equation}
\L_{depth}^s = \min\left(\L_{depth}^{\m}(\nv\D^s, \D),\L_{depth}^{\M^*}(\nv\D^s, \D)\right),
\end{equation}
We also use the multi-scale depth smoothness  regularization~\cite{godard2017monodepth} for both full-rendered $\nv\D$ and static depth $\nv\D^s$.

\topic{Plane transparency smoothness.} Relying solely on smoothing composited depths is insufficient to smooth the geometry in 3D space. Therefore, we further apply a total variation loss $L_{\a}^{tv}$ to each warped plane transparency $\nv{\a}^{\Delta^\v}_i$.

\topic{Scene flow regularization.} Depth estimation for dynamic content in a monocular video is an ill-posed problem. 
Many existing methods first estimate depth for individual frames with a single-image depth estimator. They then assume that the motion is slow and accordingly regularize the scene flows to smooth the individually estimated depth maps to improve the temporal consistency.
However, it still highly depends on the initial single depth estimates. We observe that the assumption may not always hold true and may 
compromise the scene reconstruction and view synthesis quality.
In addition, scene flow regularization slows our training process significantly. Hence, we disable scene flow regularization by default.
We discuss its effect in detail in the supplementary material. 


\topic{Implementation details.} Our implementation is based on PyTorch with Adam~\cite{adam} and VectorAdam~\cite{vectoradam} optimizers along with a gradient scaler~\cite{philip2023floaters} to prevent floaters. We follow 3D Gaussians~\cite{3dgaussian} to gradually increase the number of bands in SH coefficients during optimization. The optimization only takes 15 minutes on a single A100 GPU with 2000 iterations. We describe our detailed loss terms and hyper-parameter settings in the supplementary material.

%% file: figures/fig3_method_overview.tex
\begin{figure}[tb]
    \centering
    \includegraphics[trim={3cm 3.95cm 3.5cm 2.6cm},clip,width=\linewidth]
    {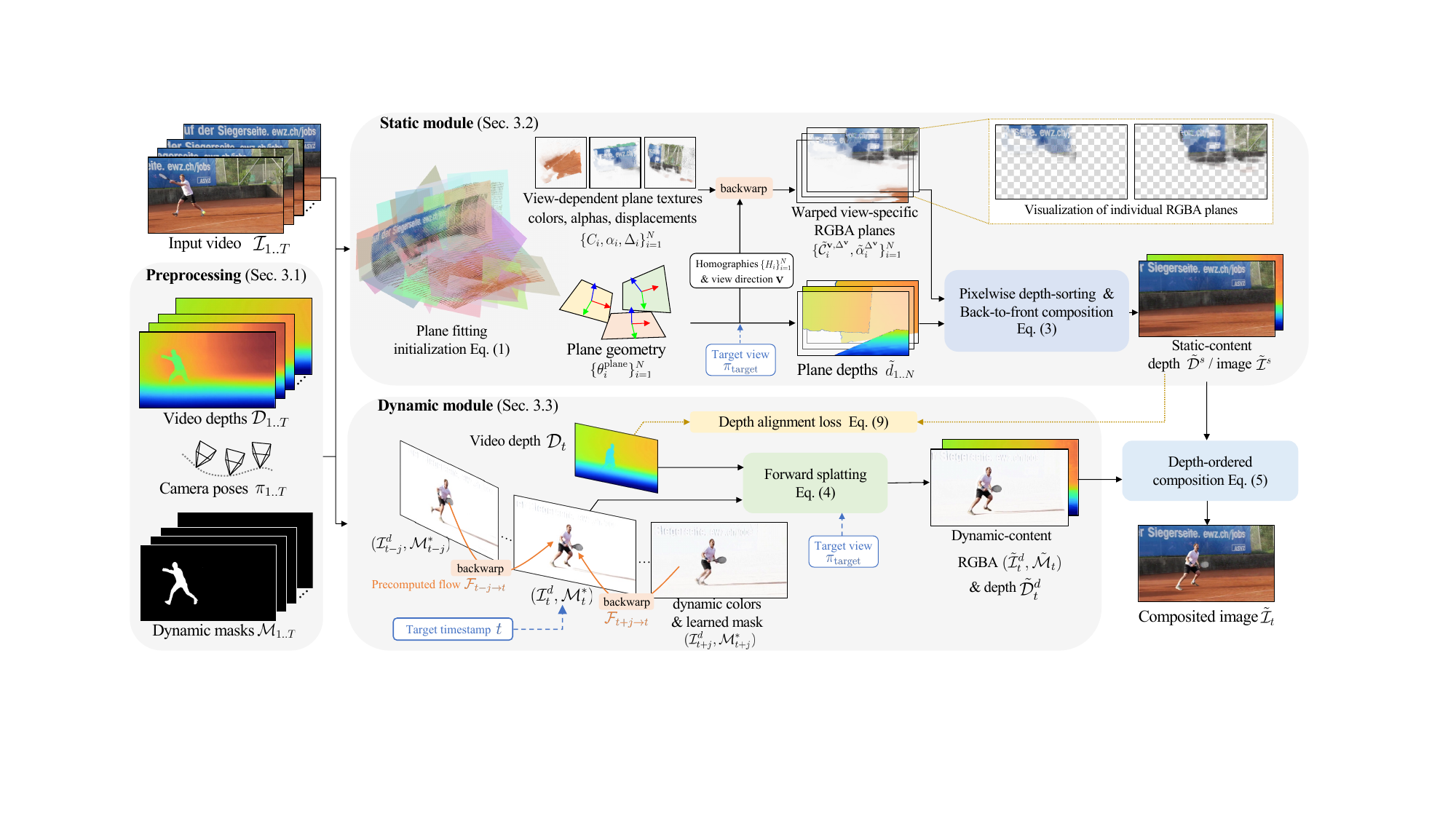}

\setlength{\abovecaptionskip}{0cm}
\setlength{\belowcaptionskip}{-0.2cm}

\caption{\textbf{Method overview.} 
We first preprocess an input monocular video to obtain the video depth and pose as well as the dynamic masks (Sec.~\ref{sec:preprocessing}). 
The input video is then decomposed into static and dynamic content. We initialize a soup of oriented planes by fitting them to the static scene. These planes are augmented to capture view-dependent effects and non-planar complex surfaces. These planes are warped to the target view and composited from far to near to generate the target static view (Sec.~\ref{sec:static_model}). We estimate per-frame point clouds for dynamic content together with dynamic masks (Sec.~\ref{sec:dynamic_model}). For temporal consistency, we use optical flows to blend the dynamic content from neighboring frames. The blended dynamics is then warped to the target view. Finally, the target novel view is composited by the static and dynamic content.
}
\label{fig:method_overview}
\end{figure}

%% file: figures/fig4_view-dependent_texture.tex
\begin{figure}
\centering
\includegraphics[trim={11.35cm 10cm 3.6cm 3.1cm},clip,width=0.98\linewidth]
{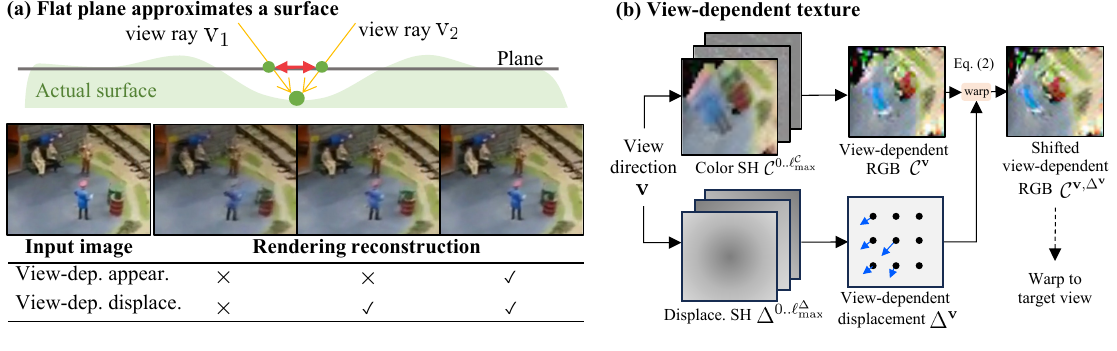}

\setlength{\abovecaptionskip}{0cm}
\setlength{\belowcaptionskip}{-0.2cm}

\caption{\textbf{View-dependent texture.} 
(a) Since a flat plane cannot sufficiently represent a non-flat surface, different viewing rays look at the same actual point but hit the plane in different locations (\textcolor{red}{red arrow}) and query different RGBA values.
(b) We augment it with both view-dependent color and displacement. The RGBA should be displaced to different locations depending on different viewing rays. Both of them are encoded by spherical harmonic coefficients, $\C^{0..{\ell_{\max}^{\C}}}$ and $\Delta^{0..{\ell_{\max}^{\Delta}}}$, respectively. Given a view direction $\v$, we first obtain the view-specific color $\C^\v$ and displacement $\Delta^\v$, then displace $\C^\v$ into the final view-specific $\C^{\v,\Delta^{\v}}$ texture for planar homography warping to the target view. Note that the transparency map $\a$ is shifted jointly with $\C^\v$.
}
\label{fig:view-dependent-texture}
\end{figure}

%% file: 4_result.tex
\section{Experimental Results}
\label{sec:result}
\subsection{Comparisons on the NVIDIA Dataset}
The NVIDIA's Dynamic Scene dataset~\cite{yoon2020nvidia} contains nine scenes simultaneously captured by 12 cameras on a static camera rig. 
To simulate a monocular input video with a moving camera, we follow the protocol in DynNeRF~\cite{gao2021dynerf} to pick a non-repeating camera view for each timestamp to form a 12-frame video. 
We measure the PSNR and LPIPS~\cite{lpips} scores on the novel views from the viewpoint of the first camera but at varying timestamps. 
We show the visual comparisons in Fig.~\ref{fig:nvidia_visual} and an overall speed-quality comparison in Table.~\ref{tbl:speed}. We provide visual comparisons with HyperNeRF~\cite{hypernerf} and 4D-GS~\cite{wu4dgaussians} and a table of per-sequence PSNR and LPIPS scores in the supplementary material. 
Overall, our method achieves the second-best LPIPS but with over 100\texttimes\ faster training and rendering speeds than NSFF~\cite{nsff}, DynNeRF~\cite{gao2021dynerf}, and RoDynRF~\cite{robustdynrf}.
\input{figures/fig5_visual_comp_nvidia}
\input{figures/fig6_comp_nvidia_long_all}

To compare with DynIBaR~\cite{dynibar}, we followed their protocol to form longer input videos with 96-204 frames by repeatedly sampling from the 12 cameras~\cite{dynibar}. 
The overall quantitative scores and time comparisons are presented in Fig.~\ref{fig:nvidia_long_all}.
Due to the ill-posed dynamic video depth without scene flow regularization, our method produces slight misalignments to the ground truth in dynamic areas. 
Nevertheless, our approach provides sharper and richer details in dynamic content than NSFF~\cite{nsff} and DynNeRF~\cite{gao2021dynerf}, and accordingly, our method achieves the second-best perceptual LPIPS score.


\subsection{Visual Comparisons on the DAVIS Datatset} 
The videos generated from the NVIDIA dataset~\cite{gao2021dynerf} are different from an in-the-wild video. 
Thus, we select several videos from the DAVIS dataset~\cite{davis} to validate our algorithm in real-world scenarios. For a fair comparison, we use the same video depth and pose estimation from our preprocessing step for DynIBaR and then run DynIBaR with their officially released code~\cite{dynibar}. 
In Fig.~\ref{fig:davis_visual}, although DynIBaR can synthesize better details with neural rendering in the first example, it introduces blurriness by aggregating information from local frames and yields noticeable artifacts in the other four examples. In contrast, our method maintains a static scene representation and obtains comparable quality to DynIBaR in the first example while being significantly faster to train and render.
\input{figures/fig7_visual_comp_dynibar}
\input{tables/tb3_speed}
\subsection{Speed Comparison}
We compare both per-scene training and rendering speed in Table~\ref{tbl:speed} along with the overall LPIPS scores on the NVIDIA dataset~\cite{yoon2020nvidia} with two protocols~\cite{gao2021dynerf,dynibar}. 
Notably, most methods require SfM preprocessing to derive camera poses and/or video depth. Such preprocessing, including COLMAP~\cite{colmap} and video depth and pose estimation methods~\cite{zhang2021consistent,robustcvd,casualsam} may require 0.5 to over 3 hours of computation. Therefore, our speed comparison primarily focuses on the main training process of scene representation and the rendering time during inference.

NeRF-based methods~\cite{nsff,gao2021dynerf,robustdynrf} usually demand multiple GPUs and/or more than a day for per-video optimization. While some recent studies~\cite{busching2023flowibr,tian2023mononerf,zhao2024pgdvs} attempt to develop a generalized NeRF-based approach, there still exists a quality gap compared to per-video optimization methods, and their rendering speed is still slow.
RoDynRF~\cite{robustdynrf}, although capable of operating without SfM preprocessing, reports an LPIPS score of 0.065 when using COLMAP preprocessing for equitable comparison, whereas the score without COLMAP stands at 0.079. Notably, their main training time is still 37 times longer than our entire process, including preprocessing (0.5 hr) and our main training process (0.25 hrs). 

In contrast to NeRF approaches, explicit representations, such as our method and 4D-GS~\cite{wu4dgaussians}, can train and render fast. 
Our method has a similar rendering speed but is faster to optimize than 4D-GS since it trains a deformable MLP. 
In summary, our method can generate novel views comparable to NeRF methods while being markedly faster to train and render (\textgreater100\texttimes\ faster than~\cite{dynibar}). 

\subsection{Ablation Study}
To thoroughly examine our method, we conduct ablation studies on the NVIDIA dataset in Table~\ref{tbl:ablation}. 
In our static module, the view dependency of plane textures plays a key role in rendering quality. The individual view-dependent appearance 
and displacement 
can each enhance the LPIPS scores by 19\% {\small(0.104\textrightarrow0.084)} and 16\% {\small(0.104\textrightarrow0.087)}, respectively. Jointly, they can improve the baseline further, 22\% in LPIPS and 1.4dB in PSNR.
The improvement by adding neighboring blending is not significant for the dynamic module since the preprocessed masks are already good as provided by DynNeRF's protocol~\cite{gao2021dynerf}. 
We demonstrate the improved temporal consistency of casual videos on our project page. 
\input{tables/tb4_ablation}

\input{figures/fig8_in-the-wild_results}
\subsection{Novel view results on in-the-wild videos}

Given that DAVIS videos~\cite{davis} often exhibit limited parallax effects due to primary camera rotations, we further showcase the efficacy of our approach on diverse in-the-wild scenarios, as depicted in Fig.~\ref{fig:in-the-wild-results}, with additional video results provided in our webpage. 
Furthermore, we use a different video-depth-pose preprocessing method~\cite{casualsam} and demonstrate a visual comparison with na\"ve depth warping in Fig.~\ref{fig:depth_warp} to show our robustness to different 3D preprocessing methods.
We present two examples, containing strong parallax and complex backgrounds (\eg fences and trees). 
In the second example, the preprocessed depth fails to accurately capture the detailed structure of the fence and the background building. Consequently, na\"ive depth warping introduces significant distortion, particularly noticeable in the window behind the fence. In contrast, our method mitigates the imperfect depth and handles complex structures through optimization to produce a coherent novel view result.
\input{figures/fig8.5_depth_warp}
\input{figures/fig9_limitations}
\subsection{Limitations}
\label{sec:limitations}
Our approach may fail when the preprocessed video depth and pose are severely inaccurate. In Fig.~\ref{fig:limitations}c, the static scene reconstruction is blurry because the inaccurate depth estimation leads to a poor initialization of the oriented planes. The planes may also have difficulties handling some long videos containing volumetric 360 scenes. In addition, our method cannot separate objects with subtle motion from a static background, such as videos in DyCheck~\cite{dycheck}, a challenging dataset for most existing methods. Besides, 
our method may produce an incomplete foreground (Fig.~\ref{fig:limitations}e) by forward splatting from the local source frame without a canonical dynamic template. The per-frame dynamic scene representation cannot synthesize sub-timesteps. 
We put this in our future direction to incorporate temporal frame interpolation for slow-motion synthesis.

%% file: figures/fig5_visual_comp_nvidia.tex
\begin{figure*}
    \centering
\setlength{\abovecaptionskip}{0cm}

\includegraphics[trim={0 6.5cm 2cm 4cm},clip,width=\linewidth]{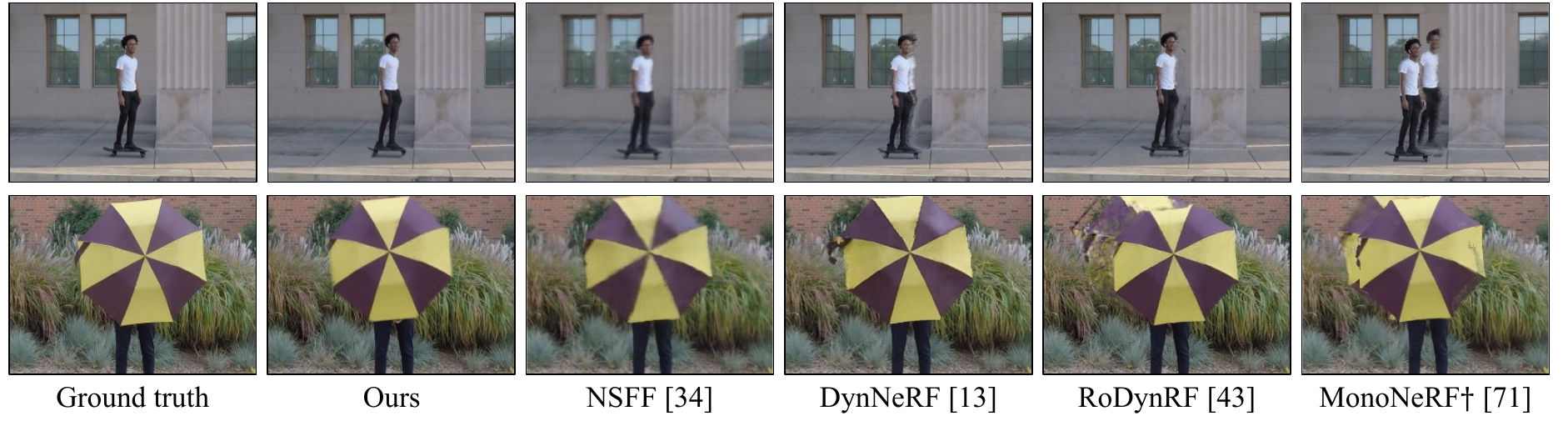}

\caption{
\textbf{Visual comparison on the NVIDIA dataset.} Our method can achieve comparable rendering quality for both static and dynamic content. Although our rendered dynamics may slightly misalign with the ground truth 
due to the ill-posed dynamic depth estimation problem, 
our results are sharp and perceptually similar to the ground truth. $\dagger$We reproduced~\cite{tian2023mononerf}'s per-scene optimization results by their official codes.
}
\label{fig:nvidia_visual}
\end{figure*}

%% file: figures/fig6_comp_nvidia_long_all.tex
\begin{figure}
\centering
\includegraphics[trim={0cm 11.7cm 1cm 0cm},clip,width=\linewidth]{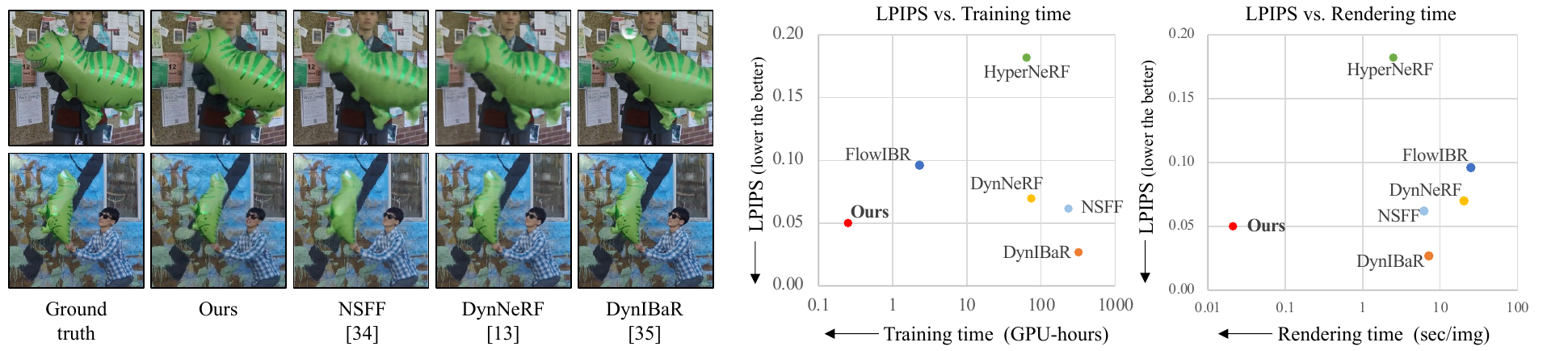}

\setlength{\abovecaptionskip}{0cm}

\caption{\textbf{Comparisons on the NVIDIA-long protocol~\cite{dynibar}.} In the visual comparison (left), 
our results are sharp with richer details than NSFF~\cite{nsff} and DynNeRF~\cite{gao2021dynerf}. Although DynIBaR~\cite{dynibar} gets the best quality overall, it is time-consuming for training and rendering (right). In contrast, with the fastest training and rendering speed, the rendering quality of our method is the second-best in the LPIPS metric.
}
\label{fig:nvidia_long_all}
\end{figure}

%% file: figures/fig7_visual_comp_dynibar.tex
\begin{figure}
\centering
\includegraphics[trim={0.7cm 5.8cm 0.4cm 7.3cm},clip,width=\linewidth]{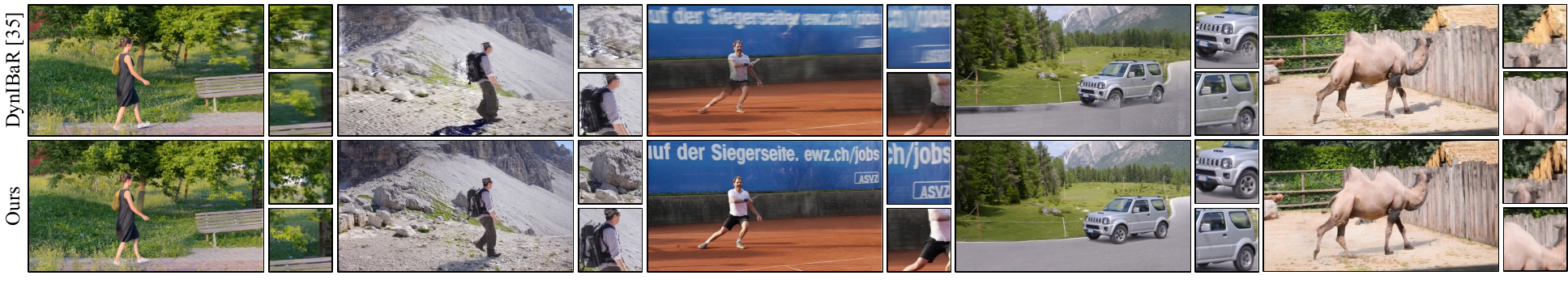}

\setlength{\abovecaptionskip}{0cm}
\setlength{\belowcaptionskip}{-0.3cm}
\caption{
\textbf{Visual comparison on DAVIS~\cite{davis}.} We showcase the novel view synthesis results and the corresponding input frame at the same time.
DynIBaR~\cite{dynibar} may fail to handle some casual videos with few parallax and introduce noticeable artifacts.
}
\label{fig:davis_visual}
\end{figure}

%% file: tables/tb3_speed.tex
\makeatletter
\newcommand*\tablethreesize{%
  \@setfontsize\tablethreesize{8}{10.0}%
}
\makeatother
\begin{table}
\caption{\textbf{Speed and quantitative quality comparison.} Our method achieves real-time rendering and the second-best LPIPS score on the NVIDIA dataset~\cite{yoon2020nvidia}. We corresponded with the authors of~\cite{yoon2020nvidia} to acquire the runtime performance. *denotes the speeds reported by~\cite{busching2023flowibr}. We highlight the best in red and the second best in yellow.}
\scriptsize
\centering
\resizebox{\textwidth}{!}{\begin{tabular}{lcr@{\extracolsep{4pt}}rr@{\extracolsep{4pt}}cc}
\toprule
\multicolumn{1}{c}{} & SfM & \multicolumn{1}{c}{Training} & \multicolumn{2}{c}{Rendering FPS} & \multicolumn{2}{c}{LPIPS\textdownarrow} \\ \cline{4-5} \cline{6-7} 
\multicolumn{1}{c}{\multirow{-2}{*}{Method}} & preprocessing & \multicolumn{1}{c}{GPU hours} & \multicolumn{1}{c}{480\texttimes 270} & \multicolumn{1}{c}{860\texttimes 480} & \cite{gao2021dynerf} protocol & \cite{dynibar} protocol \\ \midrule
Yoon~\etal~$\dagger$~\cite{yoon2020nvidia} & COLMAP~\cite{colmap} & >2 & - & <1 & 0.152 & - \\
HyperNeRF*~\cite{hypernerf} & COLMAP~\cite{colmap} & 64 & 0.400 & - & 0.367 & 0.182 \\
DynamicNeRF*~\cite{gao2021dynerf} & COLMAP~\cite{colmap} & 74 & 0.049 & - & 0.082 & 0.070 \\
NSFF*~\cite{nsff} & COLMAP~\cite{colmap} & 223 & 0.161 & - & 0.199 & 0.062 \\
RoDynRF~\cite{robustdynrf} & COLMAP~\cite{colmap} & 28 & 0.417 & 0.132 & \cellcolor[HTML]{FFCCC9}{\color[HTML]{333333} 0.065} & - \\
DynIBaR~\cite{dynibar} & Video-depth-pose~\cite{zhang2021consistent} & 320 & 0.139 & 0.045 & - & \cellcolor[HTML]{FFCCC9}0.027 \\
FlowIBR*~\cite{busching2023flowibr} & COLMAP~\cite{colmap} & 2.3 & 0.040 & - & - & 0.096 \\
MonoNeRF~\cite{tian2023mononerf} & COLMAP~\cite{colmap} & 22 & 0.047 & 0.013 & 0.106 & - \\
4D-GS~\cite{wu4dgaussians} & COLMAP~\cite{colmap} & \cellcolor[HTML]{FFFFC7}1.2 & \cellcolor[HTML]{FFFFC7}43.478 & \cellcolor[HTML]{FFCCC9}28.571 & 0.199 & - \\
Ours & Video-depth-pose~\cite{robustcvd} & \cellcolor[HTML]{FFCCC9}0.25 & \cellcolor[HTML]{FFCCC9}47.619 & \cellcolor[HTML]{FFFFC7}26.667 & \cellcolor[HTML]{FFFFC7}{\color[HTML]{333333} 0.081} & \cellcolor[HTML]{FFFFC7}0.050 \\
\bottomrule
\end{tabular}
}
\label{tbl:speed}
\end{table}


%% file: tables/tb4_ablation.tex
\begin{table}[tb]
\caption{\textbf{Ablation study.}
For static content, both view-dependent appearance $\C$ and displacement maps $\Delta$ improve the synthesis quality. For dynamics, the improvement is not significant due to the already good preprocessed masks provided by~\cite{gao2021dynerf}'s protocol. We encourage readers to view our webpage to see the improved temporal consistency.}
\centering
\tiny
\begin{subtable}{0.48\textwidth}
\centering
(a) Static scene representation
\begin{tabular}[t]{cccc}
\toprule
View-dependent & View-dependent & \multirow{2}{*}{PSNR\textuparrow} & \multirow{2}{*}{LPIPS\textdownarrow} \\
appearance & displacement &  &  \\ \midrule
\xmark & \xmark & 23.14 & 0.104 \\
\checkmark & \xmark & 24.34 & 0.084 \\
\xmark & \checkmark & 24.05 & 0.087 \\
\checkmark & \checkmark & 24.57 & 0.081 \\
\bottomrule
\end{tabular}
\end{subtable}
\begin{subtable}{0.48\textwidth}
\centering
(b) Dynamic scene representation
\begin{tabular}[t]{ccc}
\toprule
Temporal neighbor blending & PSNR\textuparrow & LPIPS\textdownarrow \\ \midrule
\xmark & 24.48 & 0.081 \\
\checkmark & 24.57 & 0.081 \\ \bottomrule
\end{tabular}
\end{subtable}
\label{tbl:ablation}
\end{table}

%% file: figures/fig8_in-the-wild_results.tex
\begin{figure}[tb]
\includegraphics[trim={0.25cm 12cm 1cm 0.25cm}, clip, width=\linewidth]{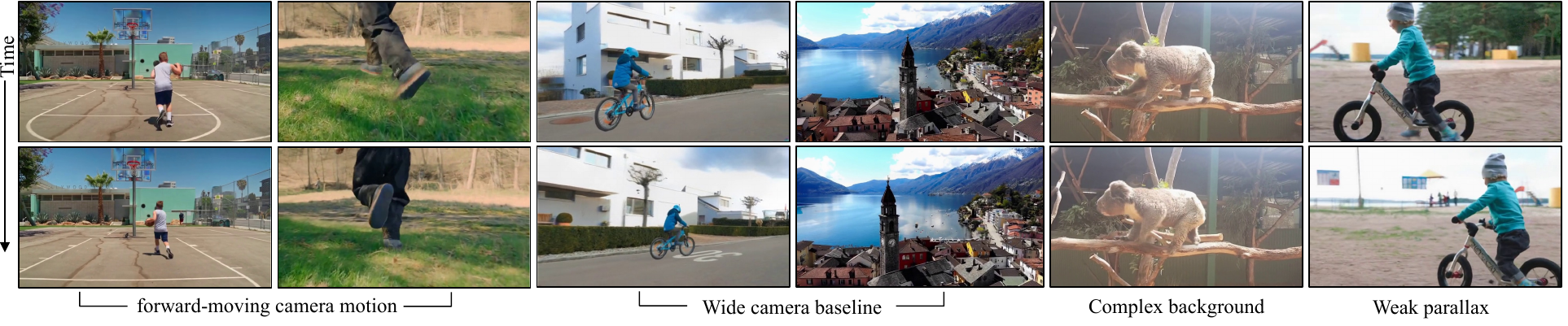}
\setlength{\abovecaptionskip}{-0.2cm}
\setlength{\belowcaptionskip}{-0.1cm}
\centering
\caption{
\textbf{Novel view synthesis results on in-the-wild videos.}
We showcase more results in diverse scenarios. Please check our project page for further video results.}
\label{fig:in-the-wild-results}
\end{figure}

%% file: figures/fig8.5_depth_warp.tex
\definecolor{mypink1}{rgb}{0.8, 0.0, 0.8}
\definecolor{myyellow1}{rgb}{1.0, 0.957, 0.0}
\begin{figure}[tb]
\centering
\includegraphics[trim={0.35cm 9.7cm 2cm 0.8cm},clip,width=\linewidth]{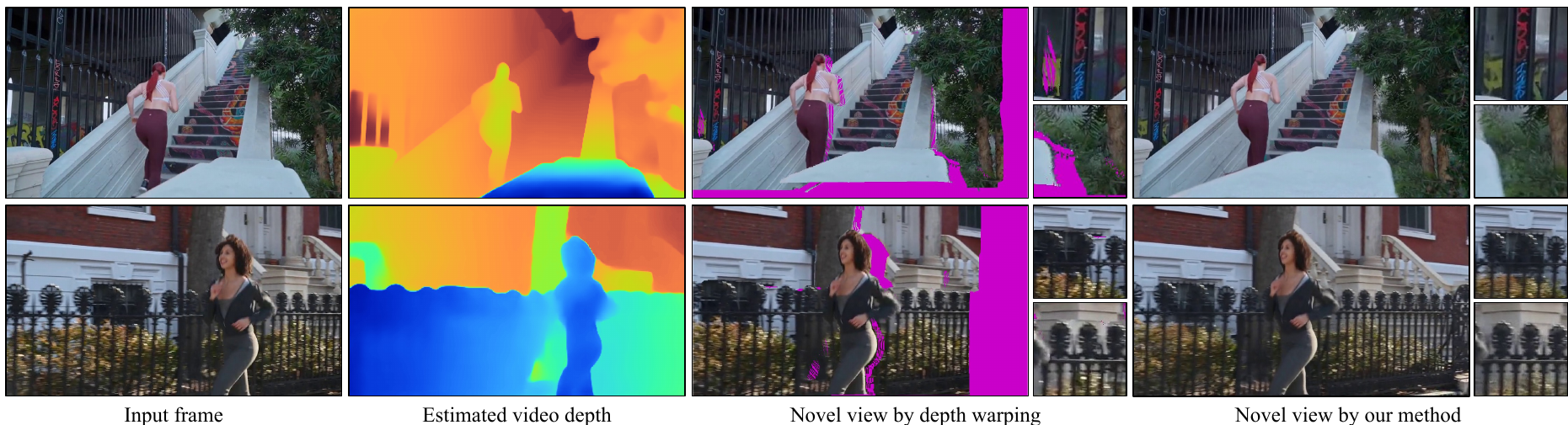} 

\caption{\textbf{Comparison with depth warping.}
We showcase two examples containing strong parallax and complex backgrounds. In the second row, the video depth behind the fence is not estimated well, causing distortions in the window behind the fence in the depth warping result.
In contrast, our method can handle the imperfect depth via scene optimization, yielding a more coherent novel view.
We highlight the \textcolor{mypink1}{unseen areas} from the input frame that are revealed in the novel view through depth warping.
}
\label{fig:depth_warp}
\end{figure}

%% file: figures/fig9_limitations.tex
\makeatletter
\newcommand*\figfailuresize{%
  \@setfontsize\figfailuresize{8}{10.0}%
}
\begin{figure}[tb]
\centering

\figfailuresize
\begin{subfigure}{0.19\linewidth}\centering\tiny
    \frame{\includegraphics[angle=90,width=\linewidth,height=1.65cm]{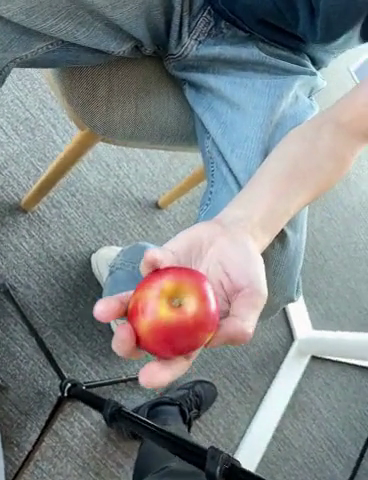}} \\
    (a) Input
\end{subfigure}
\begin{subfigure}{0.19\linewidth}\centering\tiny
    \frame{\includegraphics[angle=90,width=\linewidth,height=1.65cm]{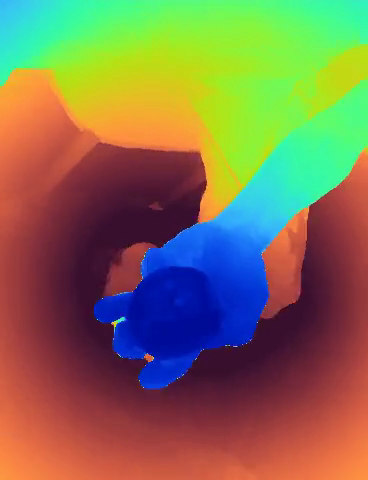}} \\
    (b) Estimated depth
\end{subfigure}\
\begin{subfigure}{0.19\linewidth}\centering\tiny
    \frame{\includegraphics[angle=90,width=\linewidth,height=1.65cm]{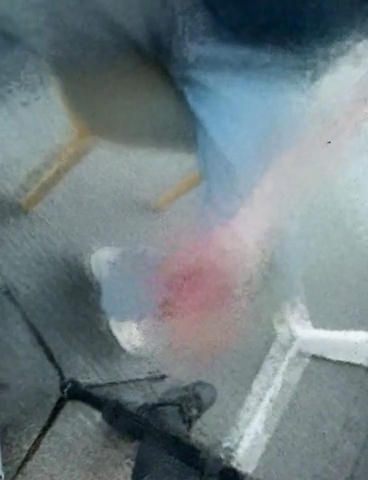}} \\
    (c) Static-only
\end{subfigure}
\begin{subfigure}{0.19\linewidth}\centering\tiny
    \frame{\includegraphics[angle=90,width=\linewidth,height=1.65cm]{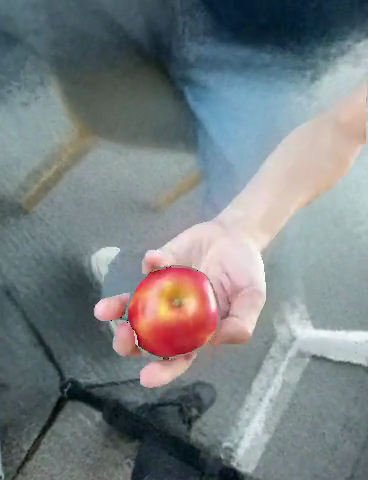}} \\
    (d) Reconstruction
\end{subfigure}
\rulesep
\begin{subfigure}{0.19\linewidth}\centering\tiny
    \frame{\includegraphics[angle=90,width=\linewidth,height=1.65cm]{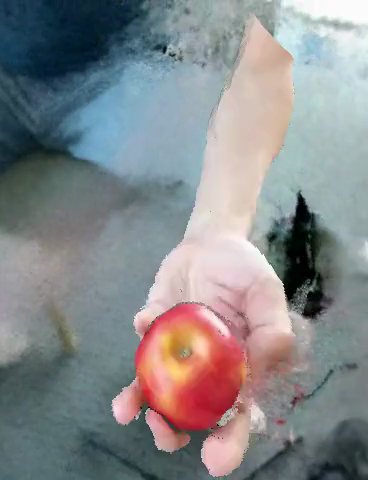}} \\
    (e) Novel view
\end{subfigure}
\caption{\textbf{Limitations.}
Our synthesis quality may degrade when the preprocessed depth (b) is severely inaccurate. The subtle dynamic motion makes motion segmentation difficult,
leaking dynamic content into the static representations (c). 
Incomplete dynamics may be revealed in distant novel views (e) due to its per-frame representation.
}
\label{fig:limitations}
\end{figure}

%% file: 5_conclusions.tex
\section{Conclusions}
\label{sec:conclusions}

This paper presents an efficient view synthesis method for casual videos. 
Similar to SOTA methods, we adopt a per-video optimization strategy to achieve high-quality novel view synthesis. 
To speed up our scene optimization process, instead of using a NeRF-based representation, we revisit explicit representations and use a hybrid static-dynamic video representation. 
We employ a soup of planes 
as a global static scene representation. 
We further augment it using spherical harmonics and displacements to model view-dependent effects and complex non-planar surface geometry. 
We use per-frame point clouds to represent dynamic content for efficiency. 
We further developed an effective optimization method together with a set of carefully designed loss functions to optimize for such a hybrid video representation from an in-the-wild video. 
Our experiments show that our method can generate high-quality novel views with comparable quality to SOTA NeRF-based approaches while being faster 
for both training and rendering. 
